\documentclass[USenglish,twocolumn]{article}
\usepackage[utf8]{inputenc}
\usepackage[big,online]{dgruyter}
\usepackage[sort&compress,square,numbers]{natbib}
\usepackage{times}
\usepackage{graphicx}
\usepackage{svg}
\usepackage{hyperref}
\usepackage{siunitx}
\begin{document}

%%%--------------------------------------------%%%
%%% Please do not alter the following lines: %%%
%%%--------------------------------------------%%%
	%\articletype{Proceedings}
  %\aop
  %\DOI{10.1515/}
  %\openaccess
  \pagenumbering{gobble}
%%%--------------------------------------------%%%

% add multi-material?
%\title{Interactive Surgical Phantom for Laparoscopic Cholecystectomy}
\title{Interactive Surgical Liver Phantom for Cholecystectomy Training}
% Oder Cholecystectomy Training?
\runningtitle{Interactive Surgical Liver Phantom for Cholecystectomy Training (as presented at CURAC 2023)}
%OpenLAP Laparoscopic Cholecystectomy Phantom ? ok schon vergeben
%OpenECP - Erlangen Cholec Phantom
%\subtitle{Insert subtitle if needed}

\author*[1]{Alexander Schüßler}
\author[2]{Rayan Younis}
\author[1]{Jamie Paik}
\author[3]{Martin Wagner}
\author[4]{Franziska Mathis-Ullrich}
\author[4]{Christian Kunz} 
\runningauthor{A. Schüßler et al.}

\affil[1]{\protect\raggedright 
  Reconfigurable Robotics Laboratory, École Polytechnique Fédérale de Lausanne, 1015 Lausanne, Switzerland, E-Mail: alexander.schuessler@epfl.ch}
\affil[2]{Department for General, Visceral and Transplantation Surgery, Heidelberg 
  University Hospital, 69120 Heidelberg, Germany} 
\affil[3]{\protect\raggedright
  Department of Visceral-, Thoracic and Vascular Surgery, Faculty of Medicine, University Hospital Carl Gustav Carus \& Center for the Tactile Internet with Human in the Loop (CeTI), Technische Universität Dresden, 01307 Dresden, Germany, E-Mail: martin.wagner@ukdd.de}
\affil[4]{\protect\raggedright 
  Friedrich-Alexander-Universität Erlangen-Nürnberg, Department of Artificial Intelligence in Biomedical Engineering (AIBE), 91052 Erlangen, Germany, E-Mail: franziska.mathis-ullrich@fau.de, christian.kunz@fau.de}

\abstract{Training and prototype development in robot-assisted surgery requires appropriate and safe environments for the execution of surgical procedures.
Current dry lab laparoscopy phantoms often lack the ability to mimic complex, interactive surgical tasks. 
This work presents an interactive surgical phantom for the cholecystectomy. 
The phantom enables the removal of the gallbladder during cholecystectomy by allowing manipulations and cutting interactions with the synthetic tissue. The force-displacement behavior of the gallbladder is modelled based on retraction demonstrations.
The force model is compared to the force model of ex-vivo porcine gallbladders and evaluated on its ability to estimate retraction forces.}


\maketitle

\section{Introduction}
With the rise of robot-assisted surgery (RAS), there is an increasing need of safe environments for the replication of surgical procedures.
While telemanipulation based RAS requires extensive training for surgeons to learn complex surgical skills without tactile feedback, machine learning based algorithms for autonomous task execution on RAS systems require extensive data collection to learn specific surgical tasks \cite{1, 2}.
%\cite{Sridhar2017, Haidegger2022}.

\begin{figure}[htb]
        \includegraphics[width=\linewidth]{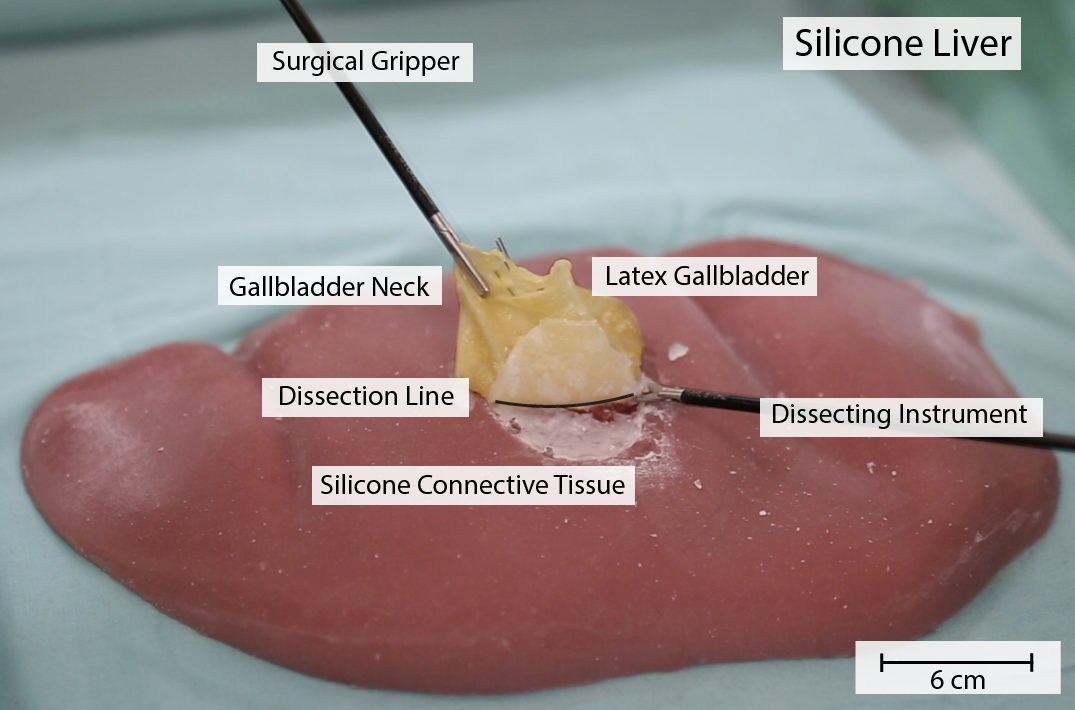}
        \caption{\textbf{Phantom}: Overview of the interactive phantom during a removal procedure. The surgical gripper for grasping the gallbladder and the dissecting instrument for the removal of the gallbladder are visible.}
        \label{fig:phantom}
\end{figure}

Training of surgical procedures are usually performed in virtual reality simulations, dry lab training environments or wet lab training environments with real tissue \cite{1}.
A recent study by Wang et al. \cite{3} shows that dry lab training on the actual robotic system results in significantly higher muscle activation and path length compared to training in virtual reality.
The study indicates that virtual simulations are cost-effective, scalable and safe, but the absence of the physical system might over-simplify the training tasks, which could lead to ineffective training of the surgeon.
However, dry lab training platforms are currently limited to the training of basic surgical skills such as cutting, suturing, and grasping.
More complex surgical tasks that require tissue handling and understanding of the tissue-instrument interaction are currently performed in wet lab environments \cite{1}.
The interactive surgical phantom presented in this work tries to increase the capabilities of dry lab training phantoms in order to close the gap between dry and wet lab training environments.

Several laparoscopic training platforms exist.
The Heidelberg laparoscopy phantom (OpenHELP) mimics the abdomen of a male patient with realistic anatomy.
While the phantom offers the possibility to grasp organs with laparoscopic instruments, it does not enable a realistic retraction and dissection task for the cholecystectomy \cite{4}.
Wang et al. \cite{5} propose a laparoscopic training platform based on ex-vivo tissue that provides a realistic experience of the laparoscopic cholecystectomy.
Dayan et al. \cite{6} and Ulrich et al. \cite{7} developed low-cost laparoscopic training platforms to practice basic laparoscopic skills, such as rope passing, peg transfer, and knot tying.
The Lübeck Toolbox trainer presents a commercially available laparoscopic video trainer for learning basic skills in minimally invasive surgery, such as peg transfer, cutting and suturing \cite{8}.

However, the integration of interactive features in surgical phantoms for the simulation of complex surgical tasks remains a challenge in dry lab environments.
Interaction force modeling provides surgeons with valuable information about the behavior of synthetic tissue compared to the behavior of real tissue.
However, the modeling and evaluation of accurate force interactions is often difficult to realize.

In this work, we present a liver with attached gallbladder phantom for the realistic execution of the gallbladder dissection step in cholecystectomy (as depicted in Fig. \ref{fig:phantom}).
The main contributions of our work can be summarized as:

\begin{itemize}
     \item A novel phantom design with interactive features to reproduce the retraction and dissection of the gallbladder during the cholecystectomy.
     %after clipping and cutting during the cholecystectomy.
     \item Non-linear force model for the gallbladder retraction of the phantom and its comparison to the gallbladder retraction on an ex-vivo porcine liver.
     \item Experimental validation of the phantom and the presented force model.
\end{itemize}

\section{Materials \& Methods}
The following chapter gives an overview of the design and manufacturing process used for the presented phantom.
A force model for the manipulation of the gallbladder and its experimental validation are presented.
\subsection{Phantom Design}
\begin{figure}
        \includegraphics[width=\linewidth]{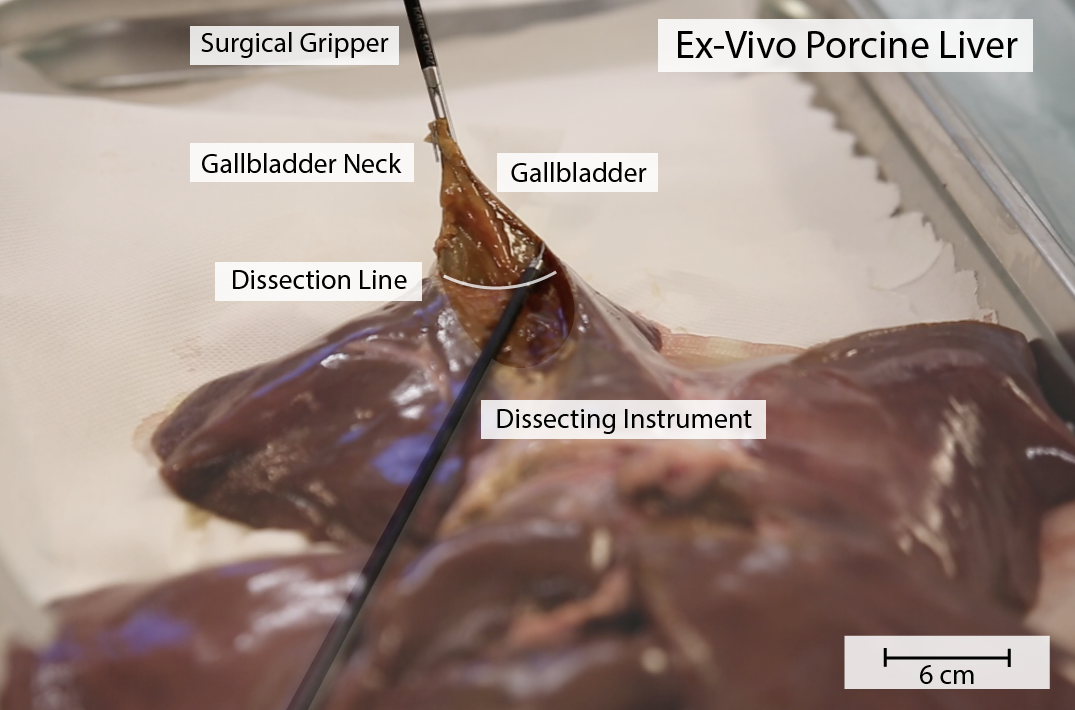}
        \caption{\textbf{Ex-vivo porcine liver:} Overview of an ex-vivo porcine liver during cholecystectomy.}
        \label{fig:exvivo}
\end{figure}

The goal of the interactive phantom is the reproduction of the retraction and dissection process during the cholecystectomy.
The interactive liver phantom presented in Fig. \ref{fig:phantom} is designed as an artificial replica of an ex-vivo porcine liver with attached gallbladder (Fig. \ref{fig:exvivo}).

The phantom consists of three main parts: A silicone liver, silicone connective tissue, and a gallbladder made from latex.
The latex gallbladder is connected to the silicone liver through the silicone connective tissue layer, which replicates the attachment of the gallbladder to the liver bed.
Similar to the real gallbladder, the latex gallbladder can be filled with liquid and closed, e.g. with staples or glue.
The dissection of the gallbladder from the liver bed during the cholecystectomy is emulated by severing the silicone connective tissue.

For safe dissection of the gallbladder from the liver, tension at the dissection line of the gallbladder is necessary.
The tension is realized through a retraction movement, which can be achieved by grasping and manipulating the latex gallbladder of the liver phantom. 

\subsection{Manufacturing Process}
\begin{figure*}[htb]
\centering
\includegraphics[width=\textwidth]{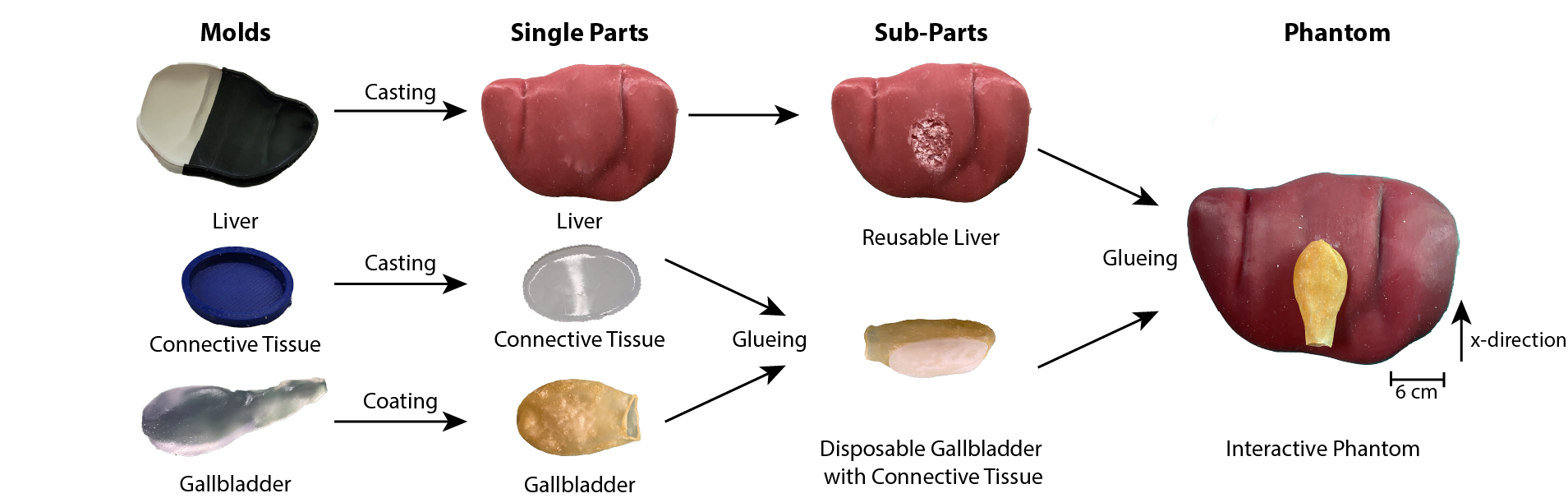}
\caption{\textbf{Manufacturing process:} The multi-step manufacturing process of the interactive phantom from manufacturing the molds and single parts to assembling the complete phantom.}
\label{fig:manufacturing}
\end{figure*}

The interactive liver phantom is manufactured in a multi-step process (Fig. \ref{fig:manufacturing}).
The basis of the manufacturing process are the negative mold of the liver, the negative mold of the connective tissue and the positive mold of the gallbladder.
The molds of the liver and connective tissue are 3D printed from PLA material using Fused Deposition Modelling (FDM).
The positive gallbladder mold is manufactured by using Stereolithography (SLA) of clear resin (Formlabs GmbH, Germany) material.
In the subsequent step, the molds are used to cast the silicone liver and silicone connective tissue with Ecoflex 00-30 silicone (Smooth-On Inc., USA).
The silicone used for the liver is dyed with red color pigments (Smooth-On Inc., USA).

Multiple layers of latex are coated and dried on to the positive gallbladder form to produce the latex gallbladder.
The latex gallbladder is glued on to the silicone connective tissue by using a combination of primer (HG pro-innovations GmbH, Salzburg, Austria) and adhesive cyanoacrylate.
The combination of gallbladder and connective tissue are subsequently glued on to the liver by using the same primer for pretreatment and adhesive cyanoacrylate.
While the silicone liver can be reused for multiple surgery simulations, the silicone connective tissue and the latex gallbladder are single-use parts.

\subsection{Force Model \& Experimental Validation}
The gallbladder retraction behavior of the phantom is investigated using telemanipulated robot-assisted retraction demonstrations.
Experiments on one ex-vivo porcine liver is used for comparison of the results.
The retraction demonstrations are conducted by controlling a surgical gripper with a Franka Panda robot (Franka Emika GmbH, Germany).
The robot is equipped with a force-torque sensor (Koris Force \& Safety Components GmbH, Germany) at its wrist to collect force data during the retraction process.
The gripper position is determined using the forward kinematics of the robot.
Overall, ten retraction demonstrations on the phantom and the ex-vivo porcine liver were conducted and recorded.
All demonstrations are performed with half of the gallbladder being attached to the liver and half of the gallbladder being removed from the liver.
The initial position of the gripper is located above the dissection line.
The gallbladder neck is grasped and retracted behind the dissection line (positive x-direction in right image of Fig. \ref{fig:manufacturing}), until the necessary tension for removal at the dissection line is reached.

Similar to the force modeling for needle insertion of Okamura et al. \cite{9}, the retraction forces of the phantom and the ex-vivo porcine liver are modeled with a non-linear model.
In this case, a second-order polynomial (with $c=0$) is fitted to the average of six retraction demonstrations:
\begin{equation}
    F(x)=a*x^2 + b*x + c.
\label{eq:nonlinear_material_model}
\end{equation}
%The model was fitted to the data using the Python SciPy library.
%The parameters of the non-linear force models are shown in table \ref{tab:model_parameters}.
%\begin{table}[htb]
%\caption{\textbf{Non-linear force model:} The parameters of the force model for the tissue retraction of the simulator LACHS and the ex-vivo porcine liver.}
%\begin{tabularx}{0.5\textwidth}{lrr}
%Model Parameters 		& LACHS 			& Ex-Vivo Porcine Liver 		\\ \midrule
%a $[N/mm^{4}]$	& -5.812e+05 	& -1.455e+05 \\
%b $[N/mm^{3}]$	& 6.710e+04     & 2.974e+04 					\\
%c $[N/mm^{2}]$	& -4.357e+02 	& -1.282e+03 					\\
%d $[N/mm]$ 	& 8.295e+01 	& 2.293e+01 \\
%e $[N]$	& -5.778e-03	& 5.451e-02 					\\
%\end{tabularx}
%\label{tab:model_parameters}
%\end{table}

\section{Results}
\begin{figure}[htb]
\centering
\includegraphics[width=0.5\textwidth]{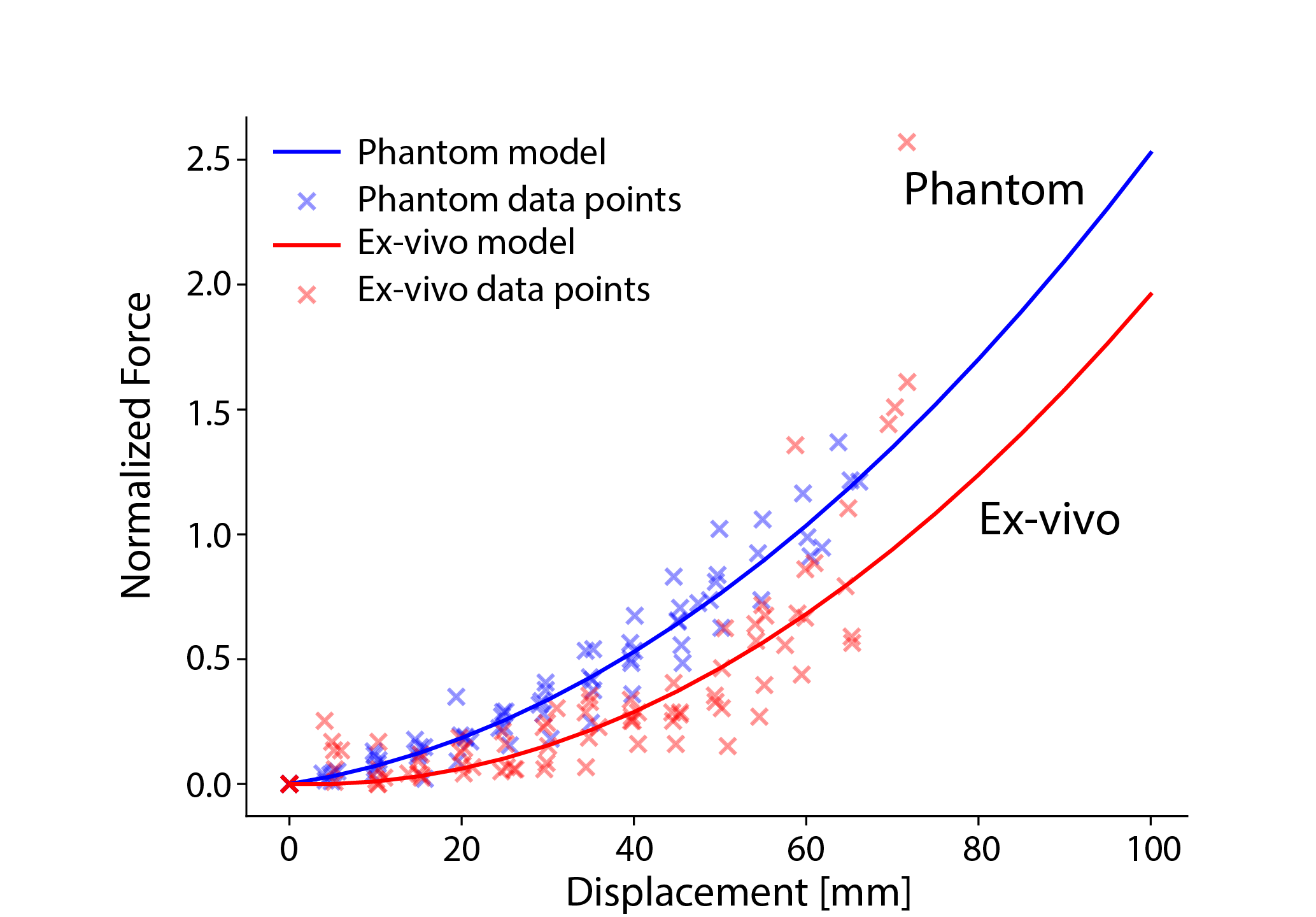}
\caption{\textbf{Non-linear force model:} Normalized force-displacement diagram with raw data points from six retraction demonstrations and force model for phantom and ex-vivo porcine liver.}
\label{fig:force_displacement}
\end{figure}

The displacement and force after reaching the desired tension at the dissection line are evaluated based on ten retraction demonstrations on the phantom and the ex-vivo porcine liver.
The retraction demonstrations are split into six demonstrations to model the retraction force behavior and four demonstrations to evaluate the model.
The force model of the phantom and of the ex-vivo porcine liver are normalized with the average retraction force (${F_{t}}_{max}$) after reaching the desired tension at the dissection line:
\begin{equation}
    F_{norm}(x)= \frac{F(x)}{{F_{t}}_{max}}.
%        F_{norm}(x)= \frac{F(x)}{F_{tension}}.
\label{eq:force_norm}
\end{equation}
The average retraction force after reaching the desired tension at the dissection line is $F_{t_{max},p}=\SI{9.97}{N}$ for the phantom and $F_{t_{max},ex}=\SI{2.06}{N}$ for the ex-vivo porcine liver.
The displacement after reaching the desired tension are $x_{t_{max},p}=\SI{56}{mm}$ for the phantom scenario and $x_{t_{max},ex}=\SI{64}{mm}$ for the ex-vivo scenario.
The normalized force-displacement diagram with data points of the demonstrations and force model for the phantom and the ex-vivo scenario are presented in Fig. \ref{fig:force_displacement}.

The presented gallbladder retraction force model is validated on four retraction demonstrations for the phantom and ex-vivo scenario.
The force model is used to estimate the retraction force after reaching the desired tension on the dissection line based on the displacement information.
The estimated force is compared to the actual retraction force of each demonstration.
The average force estimation error is $\SI[separate-uncertainty=true]{1.64(0.93)}{N}$ for the phantom and $\SI[separate-uncertainty=true]{0.16(0.10)}{N}$ for the ex-vivo scenario.

\section{Discussion}
The force model of the presented phantom gives surgeons information about the phantom's behavior during the gallbladder retraction. 
It can as well be used for the creation of a virtual simulation of the presented phantom, for usage in sim-to-real reinforcement learning methods that transfer a learned control policy from a virtual simulation to a real scenario as presented in Scheikl et al. \cite{10}.
However, there are several limitations to the presented force model.
The normalized force-displacement curves show a very similar gallbladder retraction behavior for the phantom and the ex-vivo scenario, but the average retraction force after reaching the desired tension at the dissection line used for the normalization varies between the phantom and the ex-vivo scenario (phantom: $\SI{9.97}{N}$, ex-vivo $\SI{2.06}{N}$).
The higher forces during the retraction of the latex gallbladder indicate stiffer material properties of the latex gallbladder compared to ex-vivo gallbladders.
In order to further close the gap between dry lab and wet lab surgical training, future work will investigate alternative materials to latex (e.g., polyisopren und polyurethan) for achieving retraction forces closer to the ones of ex-vivo gallbladders.
The limited retraction demonstrations for deriving the force model have a similar initial distance between the dissection line and the grasping point at the gallbladder neck, as well as the same direction of retraction.
Retraction movements to the side are not considered in the current model, but could be taken into account with a 3-DoF force model.
The force model does not consider different volumes of liquid inside the gallbladder, which may influence the retraction behavior.

\section{Conclusion}
This work presents a novel surgical phantom for the cholecystectomy.
The significance of this work lies in the interactive features of the phantom, which enable the reproduction of the gallbladder retraction and dissection process during the surgical procedure.
%The dry lab phantom provides an alternative to wet lab training based on ex-vivo tissue. 
The dry lab phantom enables more realistic training compared to current dry lab phantoms and presents an alternative to wet lab training based on ex-vivo tissue.
The retraction behavior was characterized by a force model and compared to the behavior of an ex-vivo gallbladder.

%\begin{acknowledgement}
%\end{acknowledgement}

\subsection*{Author Statement}
Research funding: This work was supported by the German Federal Ministry of Education and Research under the grant 13GW0471C and by the German Research Foundation (DFG, Deutsche Forschungsgemeinschaft) as part of Germany’s Excellence Strategy – EXC 2050/1 – Project ID 390696704 – Cluster of Excellence “Centre for Tactile Internet with Human-in-the-Loop” (CeTI) of Technische Universität Dresden.
Conflict of interest: Authors state no conflict of interest.
%\bibliography{references}
%\printbibliography

\end{document}